\documentclass[runningheads]{llncs}

\usepackage[final,year=2024,ID=21]{eccv}
\usepackage{eccv}

\usepackage{eccvabbrv}

\usepackage{graphicx}
\usepackage{booktabs}
\usepackage{tikz}
\usepackage{pgfplots}
\usepackage{wrapfig}
\usepackage{makecell}

\usepackage[accsupp]{axessibility}  

\usepackage{hyperref}
\hypersetup{pagebackref,breaklinks,colorlinks,citecolor=eccvblue}

\usepackage{orcidlink}

\begin{document}

\title{From Flexibility to Manipulation: The Slippery Slope of XAI Evaluation}



\author{Kristoffer Wickstr\o m\inst{1}\orcidlink{0000-0003-1395-7154} \and
Marina Höhne\inst{2,3,6}\orcidlink{0000−0003−3090−6279} \and
Anna Hedström\inst{2,4,5}\orcidlink{0009−0007−7431−7923}}

\authorrunning{K. Wickstr\o m et al.}

\institute{Department of Physics and Technology, UiT The Arctic University of Norway \\ \email{kwi030@uit.no} \and
UMI Lab, Leibniz Institute of Agricultural Engineering and Bioeconomy e.V. (ATB) \and 
Department of Computer Science, University of Potsdam \\ \email{MHoehne@atm-potsdam.de} \and Department of Electrical Engineering and Computer Science, TU Berlin \and Department of Artificial Intelligence, Fraunhofer HHI, Berlin, Germany \and BIFOLD - Berlin Institute for the Foundations of Learning and Data}

\maketitle

\begin{abstract}
    The lack of ground truth explanation labels is a fundamental challenge for quantitative evaluation in explainable artificial intelligence (XAI). This challenge becomes especially problematic when evaluation methods have numerous hyperparameters that must be specified by the user, as there is no ground truth to determine an optimal hyperparameter selection. It is typically not feasible to do an exhaustive search of hyperparameters so researchers typically make a normative choice based on similar studies in the literature, which provides great flexibility for the user. In this work, we illustrate how this flexibility can be exploited to manipulate the evaluation outcome. We frame this manipulation as an adversarial attack on the evaluation where seemingly innocent changes in hyperparameter setting significantly influence the evaluation outcome. We demonstrate the effectiveness of our manipulation across several datasets with large changes in evaluation outcomes across several explanation methods and models. Lastly, we propose a mitigation strategy based on ranking across hyperparameters that aims to provide robustness towards such manipulation. This work highlights the difficulty of conducting reliable XAI evaluation and emphasizes the importance of a holistic and transparent approach to evaluation in XAI. Code is available at \url{https://github.com/Wickstrom/quantitative-xai-manipulation}.
\end{abstract}

\section{Introduction}

Explainable artificial intelligence (XAI) is a crucial research area to ensure trustworthiness in computer vision~\cite{XAIbook2019}, which contains a wide range of methods that provide explanations for the output of a predictive model~\cite{Bach2015,kaihansen95,relax}. To determine which XAI method is suitable for a given problem setting, quantitative evaluation analysis is necessary to provide an objective measurement for comparison. 
Such quantitative analysis of XAI methods has made great leaps forward over the last couple of years~\cite{anna, agarwal2022openxai}, and generally consists of evaluating several metrics that measure desirable properties that an XAI method should have i.e., \textit{metric-based quality estimation}~\cite{hedstrom2023metaquantus}. However, the progress in XAI and its evaluation has led to an overwhelming variety of methods and metrics, making it challenging for researchers to navigate their choices~\cite{disagreement, hedstrom2023metaquantus, blücher2024decoupling, sanity2024}.

A fundamental limitation in XAI evaluation is the lack of ground truth explanation labels~\cite{hedstrom2023metaquantus}. Since such information is generally not available, we approximate explanation quality by measuring desirable properties like faithfulness~\cite{samek2017, ancona2019, irof, dasgupta2022, bhatt2020, road}, complexity~\cite{monotonicity, bhatt2020, complex1}, or robustness~\cite{MONTAVON20181, faithfulness-estimate, Yeh2019OnT, dasgupta2022} and translate these properties into empirical tests~. In this translation, a challenge appears in the parameterization of the empirical tests. For example, how do we mask out pixels and how large should the masks be? Preliminary works~\cite{hedstrom2023metaquantus, disagreement} have shown that the evaluation outcomes are sensitive to choices like these. This sensitivity underscores the need to investigate the impact of hyperparameter choice, making it an important research area to ensure the reliability of XAI evaluations.

This challenge becomes particularly prominent for evaluation methods with many hyperparameters that must be set, since it is generally not possible to find an objective measure of the optimal set of hyperparameters. For instance, faithfulness evaluations view model behavior changes as signals of explanations quality, with substantial changes reflecting the explanations faithfulness~\cite{Bach2015, samek2017, ancona2019, irof, dasgupta2022, bhatt2020}. This type of evaluation often requires replacing pixel values with some baseline value, which can be highly data-dependent and difficult to tune~\cite{sturmfels2020visualizing}. Furthermore, it can often be computationally impractical to evaluate all possible choices for hyperparameters. Therefore, hyperparameters are usually selected normatively with the researcher's own subjective judgment, frequently drawing on prior studies. Since there is variation in what hyperparameters are being used in the community \cite{disagreement, hedstrom2023metaquantus}, there is some flexibility in selecting hyperparameters from an acceptable set of possible choices.

\begin{table}[tb]
\parbox{.475\linewidth}{
\centering
\begin{tabular}{@{}lclll@{}}
\toprule
XAI method                       & Faithfulness score ($\downarrow$) \\ \midrule
LRP                         & 25.19                  \\
Saliency & \textbf{20.23}                  \\
Kernel SHAP                      & 23.94                  \\ \bottomrule
\end{tabular}
\vspace{0.1cm}
}
\hfill
\parbox{.475\linewidth}{
\centering
\begin{tabular}{@{}lclll@{}}
\toprule
XAI method                       & Faithfulness score ($\downarrow$) \\ \midrule
LRP & \textbf{19.31} \\
Saliency & 22.96 \\
Kernel SHAP & 24.87 \\ \bottomrule
\end{tabular}
\vspace{0.1cm}
}
\caption{Faithfulness comparison of XAI methods on MNIST before (left table) and after manipulation (right). Here, the different between the left and right table is the perturbation methods used (uniform noise vs. blurring, respectively). Both perturbation methods are commonly used, but completely change the outcome of the evaluation.}
\label{tab:motivation}
\end{table}

In this work, we demonstrate how this flexibility of XAI evaluation can be exploited to manipulate the evaluation outcome. By making seemingly small changes to hyperparameters that are widely used in the literature, the outcome of the faithfulness evaluation can change completely. \cref{tab:motivation} illustrates this, where standard XAI methods are compared with only slight changes in hyperparameters but with significant changes in evaluation outcome. We propose to frame the finding of these small changes as an optimization problem that manipulates the evaluation, where either the evaluation of a single XAI method is manipulated or the evaluation of multiple XAI methods is manipulated jointly. Our contributions are:

\begin{itemize}
    \item[\textbf{C1}] A method-specific manipulation method that can increase the evaluation score for a specific XAI method, which we entitle \textit{intra-manipulation}.
    \item[\textbf{C2}] A holistic manipulation method that can manipulate the quantitative comparison of several XAI methods, which we entitle \textit{inter-manipulation}.
    \item[\textbf{C3}] A comprehensive experimental analysis on manipulation of faithfulness evaluation that demonstrate how the evaluation outcome can completely change after manipulation.
    \item[\textbf{C4}] Towards improving the robustness of quantitative evaluation of XAI, we propose Mean Resilience Rank, a ranking-based procedure that reduces the sensitivity to hyperparameter manipulation.
\end{itemize}
Our findings have significant implications for the XAI community. Quantitative evaluation is crucial to provide objective measurements of explanation quality, which can be used to select an appropriate method for a particular task or for comparison in method development. If these measurements can be easily altered, it reduces the trustworthiness of both method selection and comparison. Therefore, the findings and solutions in this work are of critical importance for the community both by highlighting the issue of manipulation and by presenting strategies towards mitigating the issue.

\section{Related Work}

\paragraph{Metric-based Quality Estimation} Quantitative analysis of XAI explanation has improved considerably in recent years, and researchers now have a vast amount of evaluation metrics at their disposal~\cite{anna, agarwal2022openxai}. Due to the lack of ground truth explanations, researchers try to quantify the quality of an explanation by measuring desirable properties, which can be categorized into 6 families of properties~\cite{anna}; faithfulness~\cite{faithfulness-correlation}, robustness~\cite{faithfulness-estimate}, localisation~\cite{topki}, complexity~\cite{complex1}, randomisation~\cite{mprt}, and axiomatic~\cite{Kindermans2019}. Within each family, a variety of metrics exists.

\paragraph{Prior Studies on Hyperparameter Sensitivity in XAI} Increasing attention has been given to the influence and potential confounding effects of hyperparameters in XAI evaluations~\cite{hedstrom2023metaquantus}. These studies vary in defining dependent versus independent variables and the hyperparameter space of intervention, be it model, explanation, or evaluation space. Studies have examined the sensitivity of attribution methods to explanation hyperparameters like random seed and number of samples~\cite{bansal2020}, and the impact of baseline choices in methods like Integrated Gradients on explanation outcomes~\cite{sturmfels2020visualizing, integratedgradients}. Additionally, the sensitivity of explanation outcomes concerning model performance variables such as optimizer, activation function, learning rate, and dataset split has been studied~\cite{karimi2023on}, along with the effects of model priors and random weight initialization on explanations and evaluations~\cite{hase2021}. Disagreement among different explanation methods regarding top-K features and ranking has also been investigated~\cite{disagreement}, while analyzing the impact of baselines \cite{Koenen_disagree}.

Recently, researchers have explored how evaluation parameters affect outcomes, including the sensitivity of randomisation metrics to hyperparameters like normalisation, randomisation order, and similarity measures~\cite{BinCVPR23, sundararajan2018, sanity2024}. Faithfulness metrics have been examined for hyperparameter influences such as baseline choice and perturbation order~\cite{samek2017, brunke, brocki2022, rong22consistent, tomsett2020, blücher2024decoupling, barnes, DolciCGCM23}. Unlike existing work, inspired by adversarial machine learning, we introduce a novel, general-purpose manipulation approach, applicable across a variety of evaluation approaches. Our findings reveal that faithfulness evaluation outcomes are highly susceptible to manipulation. This is a key issue for the XAI community to address. We put forward a preliminary mitigating solution for this in Sec. ~\ref{sec:rank}.

\section{Preliminaries}

For clarity, we present the core concepts and notation used in the work.

\paragraph{Local explanations} Let the input to a black-box classifier $f$ be denoted as $\mathbf{x} \in \mathbb{R}^d$ and the output of the classifier as $f(\mathbf{x})=\hat{y}$. Local explanation methods~\cite{Bach2015, sundararajan2017axiomatic, guidebackprop, bykov2021noisegrad} interpret the decision of $f$ by attributing an importance score to each component of $\mathbf{x}$. We denote the explanation of $f$ for a given class $y$ as $\mathbf{e} \in \mathbb{R}^d$.

\paragraph{Evaluating Explanations} Here, we present a generalized formulation of quantitative XAI evaluation to illustrate the static input parameters and adjustable hyperparameters. We assuming an evaluation function $F \rightarrow \mathbb{R}$ on the form: 

\begin{equation}
    F(f, \mathbf{x}, \mathbf{e}, a, b, c) = s.
\end{equation}
Here, $f$, $\mathbf{x}$, and $\mathbf{e}$ are input parameters provided by the user, while $a$, $b$, and $c$ are hyperparameters that must be determined by the user. The output of the evaluation is represented by $s$, which is a scalar indicating the performance of the particular explanation. Here, we keep the hyperparameters $a$, $b$, and $c$ completely general for the sake of clarity. But note that there could be more or less hyperparameters and they can take many different forms (e.g. a number or a function), depending on the particular test and the data in questions.

\section{Manipulating XAI Evaluation}
\label{sec:manipulating}

Here, we introduce our manipulation strategies for changing the evaluation outcome of XAI evaluation with only small hyperparameter alterations. The motivation for this approach is that there often exists several agreed-upon hyperparameters for a given XAI evaluation method. For instance, when conducting a faithfulness evaluation~\cite{Bach2015, samek2017, ancona2019, irof, dasgupta2022, bhatt2020} (see \cref{sec:faithfulness} for further details), an important component is perturbing input pixels. There exist numerous methods for conducting this perturbation, and it is known that selecting a suitable one can be challenging \cite{sturmfels2020visualizing, rong22consistent, blücher2024decoupling}. However, evaluating numerous such methods can be highly computationally demanding, and due to the lack of ground truth explanations we cannot decide which method is correct. Therefore, in practice, it is common to consider only a single perturbation method \cite{MONTAVON20181, faithfulness-estimate, faithfulness-correlation}. However, as we have shown in \cref{tab:motivation}, even a slight change in the hyperparameter setting can have a big impact on the evaluation. Those who are aware of this sensitivity can potentially exploit it, which is the motivation for our manipulation strategy.

\paragraph{Intra-manipulation} 
We propose two ways to manipulate XAI evaluation methods. First, we propose to focus on manipulating the evaluation outcome for a single XAI method, which we refer to as \textit{intra-manipulation} and is defined as:

\begin{definition}[Intra-Manipulation]\label{def:intra}
  Given an evaluation function $F$, an input sample $\mathbf{x}$, an explanation $\mathbf{e}$, hyperparameters $a$, $b$, and $c$, and a feasible set of hyperparameters $A^{*}_a$ for the hyperparameter $a$, the intra-manipulation method solves the following optimization problem to determine the hyperparameter $a$, which maximizes the evaluation score of $F$:
    \begin{equation*}
        \begin{aligned}
        & \underset{a}{\text{maximize}}
        & & F(f, \mathbf{x}, \mathbf{e}, a, b, c) \\
        & \text{subject to}
        & & a \in A^{*}_a.
        \end{aligned}
    \end{equation*}
\end{definition}
\cref{def:intra} defines an optimization problem where the goal is to find hyperparameters that maximize the evaluation outcome, but are constrained to lie within a feasible set of values ($A^{*}_a$ in this case) for the hyperparameters in questions. Determining this feasible set requires a researcher's judgment and a good understanding of the particular XAI evaluation method that the user wants to manipulate. But more deeply, it fundamentally depends on the model: i.e. the feasible set is and should be dependent on the learned functional response of the model. In \cref{sec:faithfulness}, we further explain how to determine the feasible set. If the feasible set is large, \cref{def:intra} can be solved through suitable optimization techniques. If the feasible set if small, an exhaustive search can be performed. Also note that \cref{def:intra} can be extended to optimize across several hyperparameters, e.g. maximizing both $a$ and $b$.

\paragraph{Inter-manipulation} 
\cref{def:intra} allows for improving the evaluation outcome of a single XAI method. But in many cases it could be desirable to alter the outcome of the evaluation of several XAI methods. Our second manipulation approach is to take a holistic view and manipulate the evaluation of several XAI methods jointly. We refer to this approach as \textit{inter-manipulation} and define it as:

\begin{definition}[Inter-Manipulation]\label{def:inter}
  Given an evaluation function $F$, an input sample $\mathbf{x}$, a set of explanations \{$\mathbf{e}_1, \cdots$, $\mathbf{e}_M$\} from $M$ different XAI methods, hyperparameters $a$, $b$, and $c$, and a feasible set of hyperparameters $A^{*}_a$ for the hyperparameter $a$, the inter-manipulation method solves the following optimization problem to determine the hyperparameter $a$, which maximizes the following objective:
    \begin{equation*}
        \begin{aligned}
        & \underset{a}{\text{maximize}}
        & & F(f, \mathbf{x}, \mathbf{e}_m, a, b, c) - \sum\limits_{m' \neq m}F(f, \mathbf{x}, \mathbf{e}_{m'}, a, b, c) \\
        & \text{subject to}
        & & a \in A^{*}_a
        \end{aligned}.
    \end{equation*}
\end{definition}
Here, $\mathbf{e}_m$ is the explanation from the XAI method we wish to improve the performance of. We entitled this method \textit{the focus method}. The explanation from a \textit{non-focus method} is denoted as $\mathbf{e}_m'$, which we seek to worsen the performance of. The optimization problem presented in \cref{def:inter} is more complex compared to \cref{def:intra} due to the interplay between the different XAI methods. For example, the optimal solution could be found by a combination of increasing the performance of the focus-method while simultaneously decreasing the performance of the \textit{non-focus methods}. Similarly, as \cref{def:intra}, the optimization problem can be solved in several ways (e.g. Bayesian optimization) and can be extended to include several hyperparameters.

\section{Manipulating Faithfulness Evaluation} \label{sec:faithfulness}

Some types of XAI evaluation methods are more susceptible to manipulation than others. For instance, localization metrics, which aims to measure if an explanation is within a region-of-interest, usually only have 1 or even 0 hyperparameters to select \cite{topki, ARRAS202214} and are therefore harder to manipulate. On the other hand, faithfulness metrics \cite{faithfulness-correlation,faithfulness-estimate, monotonicity} have at least 3 hyperparameters that must be determined, and often more. This is one of the most popular evaluation methods in XAI~\cite{Bach2015, samek2017, ancona2019, irof, dasgupta2022, bhatt2020} and is therefore an important evaluation category to study. Therefore, we will focus on manipulating faithfulness metrics. The following section provides an overview of the fundamental components in faithfulness evaluation.

\paragraph{The fundamental components of faithfulness} Faithfulness measures to what extent explanations follow the predictive behavior of the model by iteratively perturbing the input and monitoring the corresponding change in the output of the model. Our focus will be on the task of classification, since this is the most common setting in the context of explainability and vision. This section presents the mathematical formulation of the general components of most faithfulness metrics. Let $S$ denote the set of indices $\{1,\cdots,d\}$ for each element in the input sample $\mathbf{x} \in \mathbb{R}^d$. Partition $S$ into $K$ sets $S_1, \cdots, S_K$ of equal cardinality $C$ and arranged such that:

\begin{equation}\label{eq:sorted}
    \sum\limits_{i\in S_1} e_i \geq \cdots \geq \sum\limits_{i\in S_K} e_i.
\end{equation}
For convenient notation, we define the sum of attributions for one partition as:

\begin{equation}\label{eq:atilde}
    \tilde{e}_{S_k} = \sum_{i\in S_k} e_i
\end{equation}
Inequality \eqref{eq:sorted} instructs us to rank the indices according to the input features with highest importance in a descending fashion, and are used to iteratively perturb the input. Note that some metrics sort the indices in an ascending fashion~\cite{monoarya, monotonicity, ancona2019} and some perturb the input randomly~\cite{bhatt2020}, but the general approach in faithfulness metrics is to perturb the inputs according to Equation \eqref{eq:sorted}~\cite{samek2017, rong22consistent, irof, faithfulness-estimate}. Let $\mathbf{x}_{S_1}$ denote a perturbed version of $\mathbf{x}$, where all $x_i$ for $i\in S_1$ are replaced by some baseline perturbation function $g_p$. We denote the output of the classifier based on $\mathbf{x}_{S_1}$ as $\hat{y}_{S_1}$. For $\mathbf{x}_{S_2}$, all $x_i$ for $i\in S_1 \cup S_2$ are perturbed. In general, $\mathbf{x}_{S_i}$ will have all have the indices in all sets up to set $S_i$ replaced by the baseline perturbation function.

\begin{figure}
     \centering
     \begin{subfigure}[b]{0.31\textwidth}
         \centering
         \includegraphics[width=\textwidth]{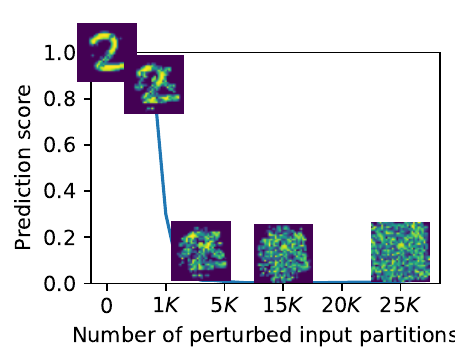}
         \caption{}
         \label{fig:faith-ex-1}
     \end{subfigure}
     \hfill
     \begin{subfigure}[b]{0.31\textwidth}
         \centering
         \includegraphics[width=\textwidth]{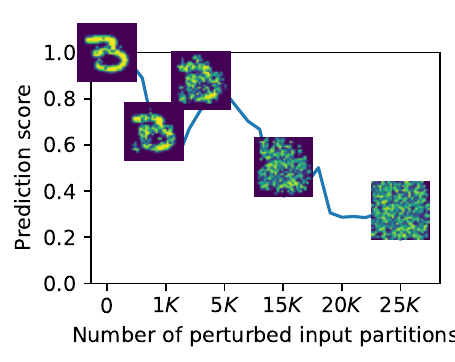}
         \caption{}
         \label{fig:faith-ex-2}
     \end{subfigure}
     \hfill
     \begin{subfigure}[b]{0.31\textwidth}
         \centering
         \includegraphics[width=\textwidth]{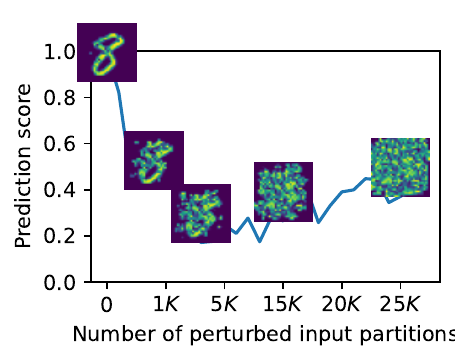}
         \caption{}
         \label{fig:faith-ex-3}
     \end{subfigure}
        \caption{Example of possible faithfulness curves for digit classification. The leftmost curve illustrates how an "intuitive" faithfulness curve might look, while the remaining curves show that there is a lot of variation in how these curves can appear.}
        \label{fig:faith-ex}
\end{figure} 

\paragraph{Illustrating the faithfulness curve} Based on the $K$ partitions of $S$, a set of progressively more perturbed inputs can be created, i.e. $\{\mathbf{x}_{S_1}, \cdots, \mathbf{x}_{S_K}\}$. Each of the perturbed inputs are classified, which gives a set of model outputs $\{\hat{y}_{S_1}, \cdots, \hat{y}_{S_K}\}$. These model outputs are the fundamental components for faithfulness evaluation in XAI. The rationale is that a good explanation should remove the essential parts of an input first, which should lead to a steep drop in the classification score. A poor explanation will remove parts that are not important, which will allow the classification score to stay high. Figure \ref{fig:faith-ex-1} shows an example where the classifier behaves as expected, with a sharp drop in accuracy when the important parts of the input are removed. To compare two explanations, one can inspect a plot such as in Figure \ref{fig:faith-ex-1} and see which explanation has the sharpest drop in classification score. However, such a visual approach has many limitations. First, we generally would like to compare explanations across many samples to get a reliable estimate of how they perform. Inspecting numerous such plots is cumbersome, and the curves can look different for different visual objects in classification, which makes comparison challenging. Also, real-world data is not always as well-behaved as the plot shown in Figure \ref{fig:faith-ex-1}, as illustrated in Figures \ref{fig:faith-ex-2} and \ref{fig:faith-ex-3}. Another important aspect is that ensuring that the curve is a genuine depiction of explanation quality and not out-of-distribution (OOD) response of the model can be highly challenging~\cite{roar, rong22consistent}.

\subsection{Hyperparameters in Faithfulness Metrics}\label{sec:hyperparams}

Here, we briefly describe the different hyperparameters that must be determined by the user to conduct a faithfulness evaluation. It is important to note that many of these hyperparameters are inherently data dependent, which means that the user must re-parameterize each metric for their use case, making the results non-comparable across different datasets and potentially models.

\paragraph{Size of partition} The size of each partition determines how many features are removed and replaced in each step of the faithfulness curve. To determine this size, there are several considerations. First, if the size of the partition is very small the evaluation will quickly become computationally infeasible, since the number of forward passes for each sample increases. Furthermore, removing only a single or a few pixels at a time can lead to adversarial effects~\cite{8601309}. Second, a large partition size will lead to course faithfulness curves which makes comparison between curves challenging. Therefore, there is a trade-off between computational efficiency and resolution of the faithfulness curves. Some researchers use the height and width of the image (assuming square images) as the size of the partition \cite{anna}, but other choices are also common \cite{relax, Bach2015}.

\paragraph{Perturbation Function} When a set of features are removed from an image, they are replaced by some perturbation function. An example of such a perturbation function could be Gaussian noise or setting pixel values to zero~\cite{faithfulness-estimate, MONTAVON20181}, but more advanced approaches are also available~\cite{road}. The type of perturbation function to apply is highly dependent on the type of images that are being considered. For example, replacing pixels with a value of zero can be possible for natural images \cite{8237633} but would not be a suitable choice for images with a black background, since this could potentially not induce a change in the network's output. In general, the choice of perturbation function varies greatly between papers \cite{MONTAVON20181, faithfulness-estimate, faithfulness-correlation, road}.

\paragraph{Aggregation Function} Examples in Figures \ref{fig:faith-ex-2} and \ref{fig:faith-ex-3}, demonstrate that it can be difficult to assess which explanation is superior. Therefore, it is desirable to aggregate the perturbed model outputs into a single score that can be easily used for comparison using an aggregation function $g_a$. There are two main approaches to aggregate the curves shown in Figure \ref{fig:faith-ex}. The first approach is to calculate the AUC of the faithfulness curve~\cite{Bach2015, samek2017, rong22consistent}. A low AUC is considered desirable, since it indicates that the important components of an input are removed first. The second approach is to correlate the model outputs with the sum of attributions within each partition~\cite{ancona2019, bhatt2020}. The motivation for this approach is that when important parts of an object are removed the predictive performance should gradually decrease, which will be captured by the correlation functions. Both correlation and AUC are used regularly in the literature \cite{Bach2015, MONTAVON20181, faithfulness-correlation, faithfulness-estimate, irof}. 

\paragraph{Normalization Function} Attributions produced by different XAI methods can have a widely different range of values. Therefore, it can be necessary to normalize the attributions such that they are comparable across different methods. A simple choice could be to standardize using the mean and standard deviation of the attributions. But choices such as these can influence evaluations~\cite{hedstrom2023metaquantus} and more sophisticated normalization schemes are also used \cite{BinCVPR23}.

\section{Towards More Reliable Quantitative Evaluation with Mean Resilience Rank}\label{sec:rank}

Due to the lack of ground truth explanations, we cannot determine what setting of hyperparameters constitutes the "correct" choice. However, we do know that it is desirable to perform well across all hyperparameter settings. Therefore, if an XAI method consistently appears among the highest-ranked methods across numerous hyperparameters, it provides an indication of high quality with less sensitivity to hyperparameters. Thus, to provide robustness towards hyperparameter manipulation, we propose to rank each XAI method for each hyperparameter setting in the feasible set, and average the ranking across the entire set. We will refer to this ranking-approach as Mean Resilience Rank (MRR).

Here, we describe mathematically how to perform this ranking. First, assume we want to evaluate $M$ explanation methods, and that we only have a single hyperparameter $a$ with a feasible set of values $A^{*}_a$ that can be altered. We denote one element of $A^{*}_a$ as $a_i$, such that the evaluation outcome for all $M$ XAI methods can be collected in the set:

\begin{equation}
    S_F(a_i) = \left\{ F(f, \mathbf{x}, \mathbf{e}_1, a_i, b, c), \cdots,  F(f, \mathbf{x}, \mathbf{e}_M, a_i, b, c)\right\}.
\end{equation}
Then, we define a function $R(\cdot)$ that takes in a set of scores and outputs a vector with integer elements, where $0$ indicates the lowest score within the set and $M-1$ indicates the highest score within the set. Finally, we define the outpout of the MMR as the following ranking vector:

\begin{equation}\label{eq:rank}
    \mathbf{r} = \frac{1}{|A^{*}_a|}\sum\limits_{a_i \in A^{*}_a} \frac{R(S_F(a_i))}{M}.
\end{equation}
For clarity, we have focused on a single hyperparameter, but \cref{eq:rank} can easily be extended to several hyperparameters. For evaluation methods where a high value is desirable, a high ranking indicates good performance, and vice versa for evaluation methods where a low value is desirable.

\section{Experimental Setup}

We evaluate our manipulation strategy across numerous datasets, models, and XAI methods, which are described below. We also define the feasible sets used in our manipulation methods.

\paragraph{Models and Datasets:} We examine several widely used computer vision datasets; MNIST~\cite{deng2012mnist}, FashionMNIST~\cite{fashionmnist}, PneumoniaMNIST~\cite{PneumoniaMNIST}, and ImageNet~\cite{deng2009imagenet}, and two common deep learning architectures: LeNet~\cite{726791} and ResNet18~\cite{resnet}. The LeNet is used for classifying MNIST, FashionMNIST, and PneumoniaMNIST, while the Resnet18 is used for classifying ImageNet. For ImageNet, we randomly sample 100 samples to conduct the faithfulness evaluation, for PneumoniaMNIST we use 500 samples, and for the remaining datasets we use 1000 samples. We choose 100 samples for ImageNet because the larger size of these images increases the computational complexity. We choose 500 for PneumoniaMNIST as it does not have 1000 samples in its test set. 

\paragraph{XAI Methods:} We investigate the following XAI methods; Layer-wise relevance propagation (LRP)~\cite{Bach2015}, Saliency~\cite{kaihansen95}, and KernelSHAP~\cite{kernelshap} using the \texttt{captum} library~\cite{captum}. We have picked these three methods as they represent common choices in the XAI field, and we have focused on only three methods to provide a clear experimental analysis without overloading the reader.

\subsection{Defining the Feasible Set of Hyperparameters for Faithfulness}\label{sec:feasible-set}

A critical aspect of the manipulation methods outlined in \cref{sec:manipulating} is to determine the feasible set of hyperparameters. This requires in-depth knowledge of the family of quantitative metrics that we aim to manipulate. In this work, we focus on the faithfulness family of evaluation metrics and the critical hyperparamters outlines in \cref{sec:hyperparams}. We focus on a subset of hyperparameters to provide a clear and understandable evaluation of our manipulation strategies. The feasible set of hyperparameters considered in this work are shown in \cref{tab:a-set}. This selection is based on common choices in the literature for partition size~\cite{Bach2015,bykov2021noisegrad,hedstrom2023metaquantus,anna,relax}, perturbation function~\cite{irof,faithfulness-estimate,sturmfels2020visualizing}, and normalization function~\cite{faithfulness-correlation,BinCVPR23,hedstrom2023metaquantus}. We consider the aggregation function fixed as AUC aggregation, which means that a lower faithfulness score is better. Specifically, we compute the AUC of the faithfulness curve from the set of perturbed model outputs $\{\hat{y}_{S_1}, \cdots, \hat{y}_{S_K}\}$.

\begin{table}[]
\resizebox{\textwidth}{!}{ 
\begin{tabular}{@{}lcccc@{}}
\toprule
                       & MNIST & FashionMNIST & PneumMNIST & ImageNet \\ \midrule
Partition size & \{14, 28, 56\} & \{14, 28, 56\}  & \{14, 28, 56\} & \{112, 224, 448\} \\
Perturbation: & \{$\mathcal{N}(0, 1)$, $\mathcal{U}(0, 1)$, $\mathcal{G}(\cdot)$\} & \{$\mathcal{N}(0, 1)$, $\mathcal{U}(0, 1)$, $\mathcal{G}(\cdot)$\} & \{$\mathcal{N}(0, 1)$, $\mathcal{U}(0, 1)$, $\mathcal{G}(\cdot)$\} & \{$\mathcal{N}(0, 1)$, $\mathcal{U}(0, 1)$, $\mathcal{G}(\cdot)$\} \\
Normalization          &    \{True, False\}   &  \{True, False\} & \{True, False\} & \{True, False\} \\ \bottomrule
\end{tabular}%
}
\vspace{0.1cm}
\caption{The feasible set of hyperparameter considered in this work for different datasets. $\mathcal{G}(\cdot)$ denotes Gaussian blurring.}
\label{tab:a-set}
\end{table}

\section{Results}

Here we present the results of performing our proposed inter-manipulation and intra-manipulation. In both cases, we survey the literature and create what we call the \textit{base} set of hyperparameters. The \textit{base} set of hyperparameters for MNIST, FashionMNIST, and PneumoniaMNIST is a partition size of 28, uniform noise as perturbations, and no normalization. For ImageNet, the \textit{base} set of hyperparameters is a partition size of 224, uniform noise as perturbations, and no normalization. After manipulation using \cref{def:intra} and \cref{def:inter}, we will obtain a new set of hyperparameters that we refer to as the \textit{manipulated} set of hyperparameters. Our results are centered around comparing the performance of the \textit{base} set and the \textit{manipulated} set.

\subsection{Intra-Manipulation Results}

\cref{tab:intra-results} shows the results of performing the intra-manipulation proposed in \cref{def:intra}, where \textit{base} is the score obtained with the selected set of hyperparameters described above and \textit{manipulated} is the score obtained after manipulation. These results demonstrate that there is much room for changing the evaluation outcome for a single XAI method, in some cases as much as a 130 \% improvement from the \textit{base} to the \textit{manipulated} evaluation outcome. Note that the \textit{manipulated} scores are not directly comparable, since the manipulation is performed method-wise and the hyperparameters can be different. Therefore, the inter-manipulation shown in the next section must be used to alter the outcome of an evaluation across methods.

\begin{table}[tb]
\centering
\begin{tabular}{lcccccccccccc}
\toprule
 & \multicolumn{2}{c}{MNIST} & & \multicolumn{2}{c}{FashionMNIST} & & \multicolumn{2}{c}{PneumMNIST} & & \multicolumn{2}{c}{ImageNet}
   \\ \cline{2-3} \cline{5-6} \cline{8-9} \cline{11-12}
XAI method & base & manip. & & base & manip. & & base & manip. & & base & manip. \\ \midrule
LRP & 25.20 & 7.86  & & 21.46 & 5.37 & & 21.31 & 6.06 & & 129.61 & 41.48 \\ 
Saliency & 20.23 & 6.80 &  & 15.65 & 4.72 & & 23.28 & 4.23 & & 124.93 & 37.53 \\ 
KernelSHAP & 23.94 & 8.01 & & 18.28 & 4.81 & & 22.06 & 4.29 & & 128.72 & 40.14 \\ 
\bottomrule
\end{tabular}%
\vspace{0.1cm}
\caption{Intra-results across several datasets and methods. Lower is better.}
\label{tab:intra-results}
\end{table}

\subsection{Inter-Manipulation Results}

\cref{tab:inter-results-lrp}, \cref{tab:inter-results-sal}, and \cref{tab:inter-results-shap} show the results of performing the inter-manipulation proposed in \cref{def:inter}, where the scores are manipulated towards LRP, Saliency, and KernelSHAP, respectively. For some tasks, the evaluation outcome can be manipulated such that most of the three methods achieves the best performance. This is particularly apparent for PneumoniaMNIST, where all XAI methods can achieve the best performance after manipulation. For some datasets there is less room for manipulation. This is most clear from the ImageNet results. That said, the evaluation difference between explanation methods can still be reduced and thus make the XAI evaluation findings less conclusive (see e.g. Imagenet results in \cref{tab:inter-results-shap}). In Appendix A, we provide a summary of the amount of times each hyperparameter occurs in the manipulated set.

\begin{table}[tb]
\centering
\begin{tabular}{lcccccccccccc}
\toprule
 & \multicolumn{2}{c}{MNIST} & & \multicolumn{2}{c}{FashionMNIST} & & \multicolumn{2}{c}{PneumMNIST} & & \multicolumn{2}{c}{ImageNet}
   \\ \cline{2-3} \cline{5-6} \cline{8-9} \cline{11-12}
XAI method & base & manip. & & base & manip. & & base & manip. & & base & manip. \\ \midrule
LRP & 25.19 & \textbf{37.79} & & 21.46 & 35.42 & & \textbf{21.31} & \textbf{43.53} & & 129.61 & 128.02 \\ 
Saliency & \textbf{20.23} & 46.23 &  & \textbf{15.65} & \textbf{34.75} & & 23.28 & 47.42 & & \textbf{124.93} & \textbf{123.93} \\ 
KernelSHAP & 23.94 & 50.77 & & 21.45 & 41.42 & & 22.06 & 45.30 & & 128.72 & 131.97 \\ 
\bottomrule
\end{tabular}%
\vspace{0.1cm}
\caption{Inter-results with manipulation towards \textit{LRP}. Lower is better.}
\label{tab:inter-results-lrp}
\end{table}

\begin{table}[tb]
\centering
\begin{tabular}{lcccccccccccc}
\toprule
 & \multicolumn{2}{c}{MNIST} & & \multicolumn{2}{c}{FashionMNIST} & & \multicolumn{2}{c}{PneumMNIST} & & \multicolumn{2}{c}{ImageNet}
   \\ \cline{2-3} \cline{5-6} \cline{8-9} \cline{11-12}
XAI method & base & manip. & & base & manip. & & base & manip. & & base & manip. \\ \midrule
LRP & 25.19 & 51.41 & & 21.46 & 43.80 & & \textbf{21.31} & 25.86 & & 129.61 & 167.14 \\ 
Saliency & \textbf{20.23} & \textbf{41.57} &  & \textbf{15.65} & \textbf{31.83} & & 23.28 & \textbf{19.61} & & \textbf{124.93} & \textbf{147.56} \\ 
KernelSHAP & 23.94 & 49.25 & & 21.45 & 37.36 & & 22.06 & 19.99 & & 128.72 & 167.74 \\ 
\bottomrule
\end{tabular}%
\vspace{0.1cm}
\caption{Inter-results with manipulation towards \textit{Saliency}. Lower is better.}
\label{tab:inter-results-sal}
\end{table}

\begin{table}[tb]
\centering
\begin{tabular}{lcccccccccccc}
\toprule
 & \multicolumn{2}{c}{MNIST} & & \multicolumn{2}{c}{FashionMNIST} & & \multicolumn{2}{c}{PneumMNIST} & & \multicolumn{2}{c}{ImageNet}
   \\ \cline{2-3} \cline{5-6} \cline{8-9} \cline{11-12}
XAI method & base & manip. & & base & manip. & & base & manip. & & base & manip. \\ \midrule
LRP & 25.19 & 12.07 & & 21.46 & 43.80 & & \textbf{21.31} & 26.42 & & 129.61 & 74.93 \\ 
Saliency & \textbf{20.23} & \textbf{9.72} &  & \textbf{15.65} & \textbf{31.83} & & 23.28 & 19.95 & & \textbf{124.93} & \textbf{74.21} \\ 
KernelSHAP & 23.94 & 11.53 & & 21.45 & 37.36 & & 22.06 & \textbf{19.55} & & 128.72 & 74.66 \\ 
\bottomrule
\end{tabular}%
\vspace{0.1cm}
\caption{Inter-results with manipulation towards \textit{KernelSHAP}. Lower is better.}
\label{tab:inter-results-shap}
\end{table}

\subsection{Towards More Robust Faithfulness Evaluation}\label{sec:ranking}

The results in \cref{tab:intra-results}, \cref{tab:inter-results-lrp}, \cref{tab:inter-results-sal}, and \cref{tab:inter-results-shap}, demonstrate that the evaluation outcome can be manipulated and can not be trusted, which reduces the trustworthiness of the quantitative evaluation. Here, we display the results of using MRR described in \cref{sec:rank} towards mitigating the potential for manipulation.

\begin{figure}
    \centering
    \includegraphics[width=\linewidth]{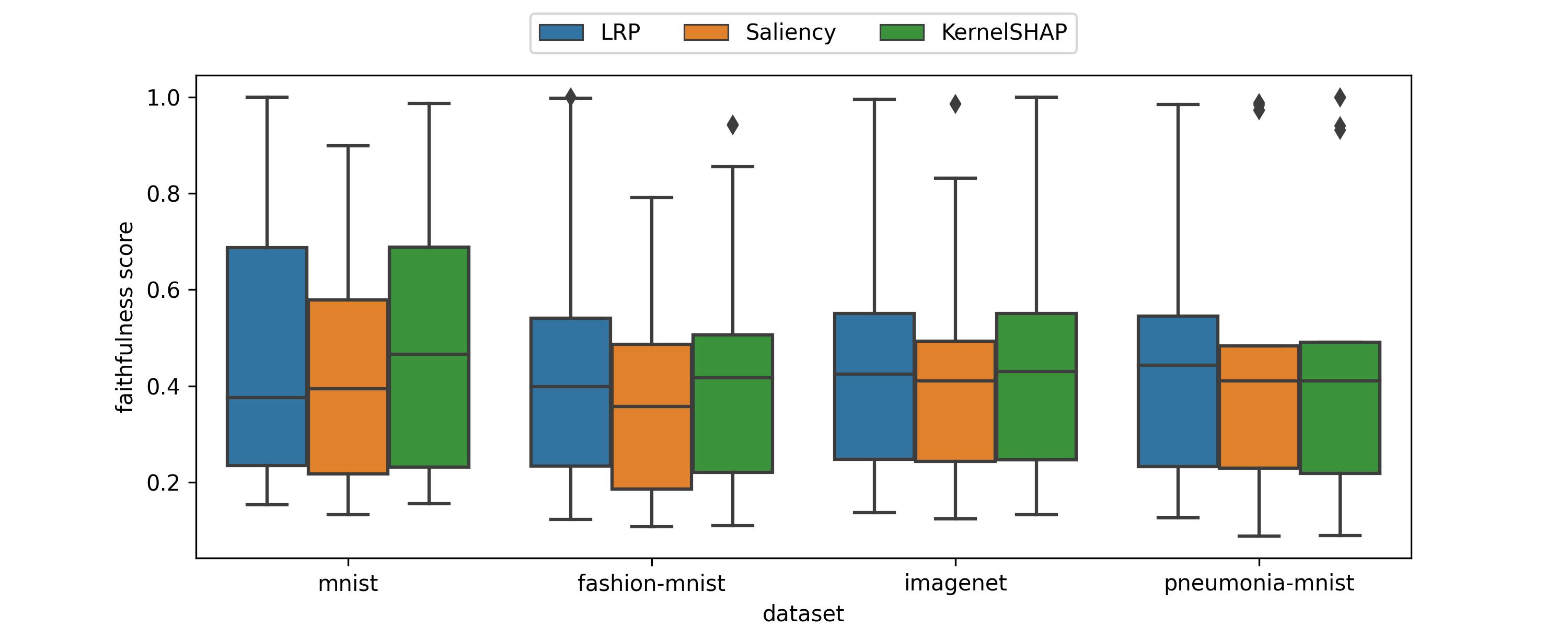}
    \caption{Box plot showing faithfulness scores across all hyperparameter configurations in the feasible set for each dataset. The plot illustrates that the average faithfulness score is similar between different XAI methods across datasets. However the high variance enables a target manipulation. Note that the scores have been normalized dataset-wise by the highest score to allow for comparison across datasets.}
    \label{fig:box-plot}
\end{figure}

\cref{tab:omni-faith} displays the results of this ranking procedure, which shows that the top-performing XAI methods change between datasets. However, if we average the ranking across all datasets, LRP comes out as the top-performing method closely followed by KernelSHAP, while Saliency seems to be consistently ranked lower. But note that there is notable variation in the scores, which we further illuminate in \cref{fig:box-plot}. The benefit of this ranking approach is that there is little room for manipulation since the top-performing methods will have to perform well across numerous hyperparameters and datasets. The downside of this ranking approach is that it requires a significant amount of computation to calculate the scores for all methods across all hyperparameters and datasets. Also, while averaging across datasets can provide robustness, it can also obfuscate important insights from a particular dataset. Therefore, it is important to include the dataset-wise ranking such that readers can get an overview of the evaluation.

\cref{fig:box-plot} shows the faithfulness score for each configuration in the feasible set for each dataset. This plot illustrates that the average faithfulness score across the feasible set can often be quite close. However, there is large spread in the scores, which is present for all datasets. This spread demonstrates the lack of robustness in the faithfulness evaluation and is part of the reason why manipulation is possible in this case. But, that alone would not be enough to allow for manipulation, since the different methods could have the same change in scores for different set of hyperparameters. However, the large standard deviation in \cref{tab:omni-faith} shows that is not the case, since the ranking change between sets of hyperparameters. In other words, the XAI methods react differently to different sets of hyperparameters. This, in combination with the variation shown in \cref{fig:box-plot}, is what allows for manipulation in this study.

\begin{table}[tb]
\centering
\begin{tabular}{lccccc}
\toprule
XAI method & MNIST & FashionMNIST & PneumMNIST & ImageNet & All \\ \midrule
LRP & \textbf{0.22} $\pm$ \textbf{0.15} & 0.33 $\pm$ 0.00 & 0.21 $\pm$ 0.00 & \textbf{0.26} $\pm$ \textbf{0.00} & \textbf{0.29} $\pm$ \textbf{0.14}  \\ 
Saliency & 0.41 $\pm$ 0.26 & 0.44 $\pm$ 0.31 & 0.37 $\pm$ 0.31 & 0.41 $\pm$ 0.33 & 0.41 $\pm$ 0.30 \\ 
KernelSHAP & 0.37 $\pm$ 0.33 & \textbf{0.22} $\pm$ 0.31 & \textbf{0.33} $\pm$ 0.27 & 0.33 $\pm$ 0.06 & 0.31 $\pm$ 0.31 \\ 
\bottomrule
\end{tabular}%
\vspace{0.1cm}
\caption{MRR across feasible set for each dataset and across datasets (last column). Lower is better, a rank of 0 is best and 1 is worst. Results show that the top performing method can change significantly between datasets, but when averaging across datasets LRP and KernelSHAP are highlighted as consistently higher ranked than Saliency.}
\label{tab:omni-faith}
\end{table}

\section{Discussion and Limitations}

The hyperparameters described in \cref{sec:feasible-set} could be extended to include other important choices such as the order of perturbation, i.e., descending or ascending~\cite{samek2017} and the type of normalization function applied~\cite{BinCVPR23}. Also, in all our experiments we repeatedly perturb the input until the entire image is perturbed, which is the standard approach in faithfulness analysis. However, when the majority of pixels are removed there is danger of OOD effects (see e.g. \cref{fig:faith-ex}), which can influence the evaluation outcome~\cite{hase2021}. An alternative approach would be to only perturb parts of an image to avoid such OOD effects. One example is to perturb until the prediction changes and then stop \cite{Bach2015}. But this introduces yet another hyperparamter, which further increases the scope for manipulation.

Our proposed MRR is a simple approach to combat the problem of manipulation, but it also has drawbacks. Most prominently, the computational cost rises quickly when more methods and hyperparameters are considered. Also, MMR requires domain expertise to determine the feasible set of hyperparameters. If the selection of the feasible set is done incorrectly, it might exacerbate the problem of manipulation since it can increase the amount of hyperparameters to choose from. MRR is also a ranking-based approach, where the scores depend on the set of explanation methods used in the analysis, including the cardinality of that set. Since the rankings are relative, they do not allow for meaningful comparisons across different tasks. To address this, we propose creating an open-source database, leveraging tools like Quantus~\cite{anna} and OpenXAI~\cite{agarwal2022openxai}, to efficiently store and standardise benchmarking results, thereby supporting researchers with the development and XAI evaluation. For future work, we further aim to expand the parameter sensitivity analysis to other families of quantitative measures such as randomisation~\cite{mprt, sanity2024} and robustness~\cite{faithfulness-estimate, Yeh2019OnT, dasgupta2022} which rely on parameters such as segmentation masks and noise perturbation methods, respectively.

\section{Conclusion}

We have presented two general-purpose methods for manipulating the quantitative evaluation of explanation methods. Intra-manipulation which increases the performance of a single method and inter-manipulation which manipulates a comparative analysis of XAI methods. The motivation for these methods is based on the lack of ground truth explanations, which makes the selection of hyperparameters in quantitative evaluation for XAI challenging. We demonstrate the effectiveness of our manipulation strategies across numerous vision datasets and XAI methods for faithfulness metrics, with results indicating that there is significant room for manipulation of the evaluation outcome. This has potentially big implications for the XAI community, as it shows that evaluation outcomes cannot always be "taken at face value" and therefore, trusted. Lastly, we present a new ranking-based procedure that aims to improve the reliability of quantitative evaluation of XAI. We believe that this work highlights the difficulty of conducting reliable XAI evaluation and emphasizes the importance of a holistic and transparent approach to evaluation in XAI.

\bibliographystyle{splncs04}
\bibliography{main}
\end{document}